\documentclass[11pt,a4paper]{article}
\usepackage[hyperref]{acl2021}
\usepackage{times}
\usepackage{latexsym}
\usepackage{multirow}
\usepackage{hyperref}

\usepackage{microtype}

\aclfinalcopy 

\title{LLMs Learn Constructions That Humans Do Not Know}

\author{Jonathan Dunn$^1$ and Mai Mohamed Eida$^2$\\
  Department of Linguistics \\
  University of Illinois Urbana-Champaign \\
  \texttt{$^1$jedunn@illinois.edu} \\
  \texttt{$^2$maimm2@illinois.edu} \\}

\date{}

\begin{document}
\maketitle
\begin{abstract}
This paper investigates false positive constructions: grammatical structures which an LLM hallucinates as distinct constructions but which human introspection does not support. Both a behavioural probing task using contextual embeddings and a meta-linguistic probing task using prompts are included, allowing us to distinguish between implicit and explicit linguistic knowledge. Both methods reveal that models do indeed hallucinate constructions. We then simulate hypothesis testing to determine what would have happened if a linguist had falsely hypothesized that these hallucinated constructions do exist. The high accuracy obtained shows that such false hypotheses would have been overwhelmingly confirmed. This suggests that construction probing methods suffer from a confirmation bias and raises the issue of what unknown and incorrect syntactic knowledge these models also possess.
\end{abstract}

\section{False Positives and Confirmation Bias}

Recent work in computational syntax has focused on the question of whether LLMs are aware of specific syntactic structures like the \textsc{let-alone} construction \citep{Bonial2024}. The goal of such work is partly to evaluate the linguistic knowledge of the models themselves but also to evaluate the learnability of these constructions without specific linguistic resources available during training. Thus, constructions which an LLM does successfully learn provide evidence for learnability, especially when these constructions are relatively rare \citep{Misra2024}.

Most previous work has followed the same high-level procedure: First, a linguist relies on their own introspection to determine that a construction exists (for them) and then annotates examples of that construction in a corpus or creates examples using their own intuitions.\footnote{The exception to this is \citealt{TayyarMadabushi2020}, which instead used a falsifiable if imperfect constructicon.} Second, these annotated examples provide stimuli for probing the linguistic knowledge of an LLM to determine whether the model is able to distinguish between this construction and other similar constructions. The procedure, in short, begins with specific constructions of interest and is limited to those constructions already hypothesized by linguists to exist for humans. For example, we start by assuming that the AANN construction exists for humans, as shown in the contrast between canonical order in (a) and non-canonical order in (b) below. Then we try to determine whether a model has also learned that construction (e.g., \citealt{Mahowald2023}).

~

\noindent\hspace{5mm}(a) five terrible weeks

\noindent\hspace{5mm}(b) a terrible five weeks \\

If the LLM is unable to distinguish or identify the construction of interest, then the conclusion is that the model is wrong in that it disagrees with the gold-standard of human introspection \citep{Weissweiler2022}. If, on the other hand, the LLM is in fact able to distinguish this construction, this is taken as evidence that the construction is learnable from usage alone \citep{Misra2024}. At no point is it possible for the introspection-driven hypothesis that a construction exists able to be disproven. And at no point is it possible to discover that the model has also incorrectly learned other constructions that humans do not know.

There are two potential issues with this line of argumentation: First, these methods are not able to discover false positives: what constructions has an LLM learned in error? Is a model \textit{aware of} constructions which humans do not know? In other words, by starting with constructions derived from introspection, these methods can only confirm or disconfirm whether that specific construction has been learned. This means that it is impossible to discover that the model has learned a new construction, where such a \textit{new construction} could be either a false positive or a construction that remains unknown to linguists or a construction from a dialect or register which linguists have yet to describe. 

This paper focuses on this problem of probing for new or false positive constructions by analyzing contextual embeddings representing sets of sentences instantiating a single construction. The goal is to find constructions which the model distinguishes as separate but which humans do not.

The second potential issue with previous approaches to construction probing is that there is often no reproducible criteria to define what actually constitutes a construction from a linguistic perspective. And yet before claiming that LLMs are in error by not knowing a construction, we would want a fully reproducible and falsifiable definition of whether some pattern does in fact constitute a construction \citep{Cappelle2024}. How can we establish, beyond personal introspection, which constructions should be known to an LLM?

This is especially problematic considering that usage-based approaches to Construction Grammar rely on a notion of \textit{entrenchment} in which some representations are more grammaticalized than others \citep{divjak_2019}. What level of entrenchment qualifies a construction as needing to be learned by an LLM and what population/register provides the baseline for measuring such entrenchment? The basic challenge is that even probing-based methods require a reproducible and falsifiable definition of what is or is not a construction, leading to a confirmation bias. Such methods can only discover that the original analysis was correct (true positives) or that the LLM is incorrect (false negatives).

\textbf{The main contribution of this paper is to ask whether LLMs make errors by over-learning constructions which do not actually exist for human speakers.}\footnote{Supplementary material is available at \href{https://doi.org/10.17605/OSF.IO/W2XYB}{https://doi.org/10.17605/OSF.IO/W2XYB}} We combine a behavioural probing task based on contextual embeddings with a meta-linguistic probing task based on prompted linguistic analysis in order to compare such false positive constructions from the perspective of both implicit and explicit linguistic knowledge.

Previous work has probed for the existence of constructions within LLMs using a variety of methods. Prompt-based approaches often rely on explicit meta-linguistic knowledge, such as asking for grammaticality judgments \citep{Mahowald2023} or providing explicit descriptions or examples of constructions \citep{Torrent2024, Bonial2024, Morin2025}. Such meta-linguistic tasks require knowledge of the language but also knowledge of linguistics; many native speakers of English, for instance, would struggle with such tasks. The experiment in Section \ref{exp1} uses this line of work to search for false positive constructions in explicit linguistic knowledge.

Other work has probed for constructions using more direct properties of models: log probabilities \citep{Hawkins2020, Leong2023} and contextual embeddings \citep{Li2022b, Weissweiler2022, Chronis2023}. Even if a model distinguishes between constructions with similar forms, however, this does not entail that the model is able to correctly interpret that difference \citep{Zhou2024}. We refer to these more direct tasks as behavioural probes in the sense that they do not require explicit linguistic analysis in the way that the prompt-based methods do. The experiment in Section \ref{exp2} uses this line of work to again search for false positive constructions, this time in implicit linguistic knowledge.

The paper is organized as follows: First, we discuss the corpus data used to probe for false positives; this consists of 100 sentences each for five clause-level constructions. Importantly, the sentences in each category are all examples of the same construction. Second, we use meta-linguistic prompts to see whether a model can be induced to mimic false analyses in a sentence sorting task. Third, we use contextual embeddings together with unsupervised methods to see whether a model distinguishes incorrectly between instances of a single construction. Together, these experiments show that LLMs hallucinate non-existing constructions and that probing experiments using these hallucinated constructions would have confirmed their existence, a significant error. These results show the need for caution in making conclusions about linguistic theory based on probing experiments.

\section{Data}

This section discusses the corpus data used to probe for false positive constructions. The basic idea is to collect 100 examples each for five separate clause-level constructions. Each of these examples should be comparable in terms of its constructional analysis, although varying in other structural and lexical and topical attributes. These sentences are collected from the Universal Dependencies English corpora \citep{Nivre2020}, chosen so that the dependency annotations can be searched for sentences which share the same form. The extracted examples are then analyzed using introspection to ensure that each set represents one and only one clause-level construction. This provides an overall corpus of 500 sentences divided into five constructional categories as described below.

The first category is derived from intransitive constructions, as shown in (1) and (2). Although the selection criteria is focused on clauses without arguments present, most of these examples contain motion-event verbs.

~

\noindent\hspace{5mm}(1) One little boy stands up.

\noindent\hspace{5mm}(2) I literally just pretty much woke up and left this morning. \\

The second category is derived from transitive constructions, as show in (3) and (4). The selection criteria is a clause with a single argument in which that argument is a noun phrase.

~

\noindent\hspace{5mm}(3) Olivia played records with the living-room windows wide open.

\noindent\hspace{5mm}(4) They just built a hotel in Syria. \\

The third category is closely related, containing transitive constructions in which the argument is an embedded clause. Examples are shown in (5) and (6). The sentences in this category have similar structures in the embedded clauses, with some degree of natural variation across them.

~

\noindent\hspace{5mm}(5) The Great Powers realized they had to change their decision.

\noindent\hspace{5mm}(6) Quinn realized that he should be going. \\

The fourth category contains single-argument clauses that have been passivized. Examples are shown in (7) and (8), both containing the original agent in a by-phrase.

~

\noindent\hspace{5mm}(7) Without a valid visa, boarding will be denied by the airline.

\noindent\hspace{5mm}(8) Tropical cyclones are sustained by a form of energy called latent heat. \\

Finally, the fifth category is double object constructions, as shown in (9) and (10). These sentences vary by whether either of the arguments are pronominal.

~

\noindent\hspace{5mm}(9) Silent, I give his case some thought.

\noindent\hspace{5mm}(10) I faxed you the promotional on the Nimitz post office. \\

As shown in these examples, the data consists of sentences with the same form and the same schematic meaning at the clausal level. While there are variations across examples of each category, in terms of lexical items and sub-clausal structures, they are examples of the same underlying clausal construction given the introspections of linguists. Our question is whether LLMs view these as coherent constructions (as humans do) or whether some instances within each category are viewed as distinct constructions (thus, false positives). In both experiments it turns out that the models do posit false positive constructions within these sets. Thus, we later conduct further introspective analysis to ensure that these distinctions are not linguistically-motivated by confounding factors.

\section{Experiment 1: Meta-Linguistic Prompts}
\label{exp1}

Our first experiment uses meta-linguistic prompts to determine whether LLMs, in this case GPT-4, hallucinate constructions that are invisible to human speakers. The basic prompting procedure is replicated from recent work on probing GPT-4 for linguistic knowledge of constructions at different levels of abstraction \citep{Bonial2024}.

This prompt-based approach is very similar to a sentence sorting task \citep{Li2022b}: the model is given the name of a construction with an example and a set of six stimuli sentences. Three of these stimuli are actual examples of the construction and the model is asked to identify them. This is the same as sorting the sentences by syntactic similarity to the example and to each other. In the original experiment, the name of the construction is drawn from the CxG literature and the examples are constructed by a linguist. For instance, the following is a possible prompt:

\begin{quote}
From amongst the following sentences, extract the three sentences which are instances of the \textsc{Let-Alone} construction, as exemplified by the following
sentence: ``None of these arguments is notably strong, let alone conclusive." Output only the three sentences in three separate lines: [\textit{Followed by six examples to sort.}]
\end{quote}

Because we are interested in discovering false positive constructions, we cannot use the name of existing constructions in the literature. First, these may be present in the training data and thus be known to the model. But, more importantly, we want to find constructions which linguists are not aware of and thus which have no name. To overcome this problem, we create five nonce construction names which could plausibly be used in linguistic description but which also do not suggest specific existing constructions: the Pristine Exemplar construction, the Reverted Focus construction, the Alternate Application construction, the Normalized Attribution construction, and the Entrenched Objective construction. These names have not previously been used and thus could hypothetically point to previously undescribed structures.

Our data consists of 100 sentences that are instances of five clause-level constructions. For each prompt, we randomly choose an example sentence from each category and six stimuli sentences. Importantly, these sentences are all instances of the same construction given the introspections of a linguist and thus the sorting task is prompting the model to create clusters of constructions, some of which match the fake construction name and example and others of which do not match. We have two goals here: First, to determine what would happen if we asked the model to undertake a spurious linguistic analysis and, second, to determine whether any new model-driven constructions are plausible (constituting unknown constructions) or hallucinations (constituting false positive constructions).

For each category in the data set we undertake 100 unique prompts for each of the construction names above; this allows us to examine whether these invented names influence the output of the model. Each of these prompts also draws on a unique example, so that we can also examine the influence of specific examples on the output.

\begin{table*}[t]
    \centering
    \begin{tabular}{|r|c|c|c|c|c|}
        \hline
         ~ &  \textit{Intransitive} & \textit{Transitive (NP)} & \textit{Transitive (C)} & \textit{Passive} & \textit{Double Object} \\
         \hline
Alternate Application & 92.69\% & 92.04\% & 92.84\% & 93.88\% & 92.74\% \\
Entrenched Objective & 93.93\% & 92.53\% & 91.70\% & 94.36\% & 91.25\% \\
Normalized Attribution & 94.53\% & 92.76\% & 92.20\% & 93.34\% & 93.79\% \\
Pristine Exemplar & 93.01\% & 93.26\% & 90.33\% & 93.18\% & 91.41\% \\
Reverted Focus & 93.39\% & 93.34\% & 93.64\% & 93.98\% & 93.81\% \\
         \hline
    \end{tabular}
    \caption{Accuracy by Percent Correct for the Consistency of Sentence Sorting by Construction Name and Across Exemplars. High Accuracy indicates that we would have confirmed an incorrect hypothesis. \textbf{These results show that the construction name has no influence on the observed sorting behaviour.}}
    \label{tab:prompt_robustness}
\end{table*}

We operationalize this question of false positives around the stability of the sentence sorting: do the same sentences end up being clustered together regardless of the artificial construction name and the provided example? If so, this means that the LLM is consistently making a distinction between sentences which are actually instances of the same construction. If the sorting were based on the invented construction name or on the randomly chosen example sentence, then the sorting patterns would vary along these two dimensions. But if, on the other hand, the sorting is based on an underlying hallucinated construction, then the name and the example would have no influence at all on the sorting. Thus, a high stability across these dimensions would mean that neither the name nor the example influence which patterns are ultimately discovered in this task.

An alternate way of viewing this experiment is as testing a hypothetical analysis: if we assume that there is in fact a construction with the given name (e.g., the Alternate Application construction) with the given example as a good instance, how would the model behave? In this hypothetical, some of the sentences are instances of this construction and others are not. We evaluate this hypothesis by looking at whether the same sentences are consistently sorted together, either as members of the positive or of the negative category. For instance, the transitive constructions in (11) and (12) are consistently grouped together four times with four separate exemplars. This would constitute 100\% agreement in the sorting. Our puppet hypothesis would thus have reached an accuracy of 100\%, confirming the validity of this hallucinated construction. The sentence in (13), on the other hand, is grouped together with (11) twice but grouped separately once. This means the model would have an accuracy of 66\% for our puppet hypothesis. Low agreement here means that there is no hallucinated construction.

~

\noindent(11) Luckily they caught the crooks before they did one on us.	\\
\noindent(12) They have good sushi for a good price. \\
\noindent(13) They can cause property damage, create a mess, and produce unpleasant smells. \\

The results across constructions and artificial construction names is shown in Table \ref{tab:prompt_robustness}. This table considers 100 random exemplars for each cell; high consistency or accuracy within a cell means that the same sorting of sentences is reached across many exemplars. The rows in the table show the invented names. Thus, if the examples influence the sorting there would be low agreement overall and if the artificial name influences the sorting there would different patterns across rows.

These results show that the sorting of sentences into two constructions is remarkably robust across both the specific exemplar given in the prompt and the name applied to the supposed construction.\footnote{Note that there are a few occasions on which GPT-4 returns ``None of the provided sentences match the X construction." And a single time is only one sentence returned as a match. Thus, this kind of response is technically possible but occurs only a few times.}

In other words, if we viewed this as an actual hypothesis, that these examples represent two distinct constructions with similar forms, these results would have confirmed our hypothesis. And yet we know that this is not a real distinction: each column represents one and only one construction, given the introspection of linguists. This is strong evidence that such a methodology has a confirmation bias: we could have confirmed any constructional analysis in this way. In short, either introspection is unreliable for identifying constructions (the linguist is wrong) or the model has hallucinated a constructional distinction which does not exist.\footnote{Because the model does not recognize the name of these nonce constructions, it could have been the case that this prompt is not specific enough. To check this, we tried alternate formulations which ensured that the analysis was linguistic in nature. For example, we added these sentences to the prompts: ``You are a linguist who is analyzing the grammar of sentences. A construction is a syntactic unit that maps between form and meaning..." However, these alternate formulations had no significant differences from the original prompt.}

Our next question is whether these new constructions which GPT-4 reliably detects are either (i) false positive hallucinations that have no linguistic regularity or (ii) meaningful constructions which were previously missed by linguistic introspection. Since the prompts reliably produce sets of sentences which the model believes represent a single construction, we use introspection to analyze some of these model-driven distinctions. To organize the data into two separate clusters, we create a vector space which captures the co-occurrence of sentences within prompt outputs; a 0 value for instance would mean that two sentences were never paired together in a response. We then use these vectors with k-means clustering to divide the sentences into two groups.

\begin{table}[t]
    \centering
    \begin{tabular}{|r|c|c|c|c|c|}
        \hline
         \textbf{Category} & \textbf{Cluster Accuracy} \\
        \hline
         \textit{Intransitive} & 74.8\% \\
         \textit{Transitive (NP)} & 71.1\% \\
         \textit{Transitive (C)} & 77.0\% \\
         \textit{Passive} & 80.5\% \\
         \textit{Double Object} & 74.2\% \\
         \hline
    \end{tabular}
    \caption{Accuracy by Percent Correct for Clusters Learned from the Sentence Sorting Task. High Accuracy means that a sentence only occurs in pairs with other sentences in the same cluster.}
    \label{tab:prompt_clusters}
\end{table}

The resulting clusters make a stronger case for hallucinated constructions: the previous analysis focused on pairs of sentences that were sorted together. Here it turns out that this pairwise relationship extends all the way to indirect groups in which sentences only occur with pairs of pairs of pairs. As shown in Table \ref{tab:prompt_clusters}, between 71.1\% and 80.5\% of sentences only occur in pairs with other sentences in the same cluster, so that these clusters explain a large portion of the sorting behaviour in this experiment. This is remarkable in that this sorting is done across many unique examples across many artificial construction names. The small-scale sentence sorting prompt produces consistent groups across many iterations, thus leaving us with these larger clusters. The next question is whether these hallucinated constructions are false positives or previously unknown structures.

~

\noindent(14a) All tropical cyclones are driven by high heat content waters. \\
\noindent(14b) As in the old days, varnish is often used as a protective film against years of dirt.\\
\noindent(14c) Sufaat was arrested in December 2001 upon his return to Malaysia. \\

An introspection-based analysis shows that there is no constructional difference between sentences in the two clusters suggested by the model. For instance, passive sentences from one cluster are given above and aligned with those from the other cluster given below. Thus, (14a) and (15a) are clearly instances of the same construction, for a human, even though they are clearly separated by GPT-4. These are examples of an hallucinated construction.

~

\noindent(15a) Pressure for change is driven by the wish of women to choose their own fate.\\
\noindent(15b) In the 21st century this book is still used as one of the basic texts in modern Structural linguistics. \\
\noindent(15c) His chief aide in Najaf was suddenly arrested along with 13 other members of his organization.\\

What does the model think these constructions look like? One clue comes from some of the very rare responses in which the sentence sorting task is not undertaken because no matches are found. As mentioned above, these occur only a handful of times. Here is one example explanation of a non-match: 

\begin{quote}
None of the sentences contain a possessive pronoun subject (like \textit{theirs}) in a subordinate clause following a verb of cognition (like \textit{knew}), with the subject of the subordinate clause being a reverted or pronominalized NP referencing a salient set from the discourse.
\end{quote}

\begin{table*}[t]
    \centering
    \begin{tabular}{|c|ccccc|}
        \hline
        ~ & \textit{Intransitive} &  \textit{Transitive (NP)} & \textit{Transitive (C)} & \textit{Passive} & \textit{Double Object} \\
        \hline
        Direct Embeddings & 0.92 & 0.88 & 0.92 & 0.91 & 0.93 \\
        Grammar-Focused & 0.85 & 0.79 & 0.87 & 0.85 & 0.85 \\
        \hline
    \end{tabular}
    \caption{Prediction accuracy (f-score) for distinguishing between clause-level constructions using both types of embeddings. A high accuracy validates that this method is able to distinguish between actual constructions. The value for each construction is the average f-score for distinguishing it from every other construction in a binary task.}
    \label{tab:validation_embeddings}
\end{table*}

This description of the example sentence is partly nonsensical but mostly far too specific to be an actual schematic construction. This provides a clue about the nature of these hallucinated constructions: they involve too specific a description over too broad a context. For instance, recent work has shown that there is a negative relationship between the size of a model and its ability to predict human reading times (in other words, showing that models with better perplexity on a test corpus make worse predictions about surprisal: \citealt{Oh2023}). The cause of this disconnect is that the model is capable of remembering infrequent patterns within very long contexts \citep{Oh2024}. Humans learn constructions precisely because they must forget the specific details of utterances and the contexts in which they occur. Constructions are remembered so that more specific details can be forgotten. 

On the other hand, these results reflect the ability of LLMs to identify and, in this case, create novel patterns; dealing with novel items is an essential part of language processing \citep{Eisenschlos2023}. The challenge here arises when this ability to create new patterns is interpreted as confirmation of the original hypothesis. This experiment would have confirmed a hypothesis that (14c) and (15c) are examples of distinct constructions.

\section{Experiment 2: Behavioural Probes}
\label{exp2}

Our second experiment uses contextual embeddings from the Pythia 1.4b model to determine whether the model is able to distinguish between two distinct constructions which have similar forms but different meanings. The main idea, however, is that these two constructions are not actually distinct. Thus, we are evaluating a false positive distinction and, if this embedding-based probe is successful in maintaining such a distinction, this is evidence for a confirmation bias. This experiment follows probing methods previously used to sort sentences \citep{Li2022b} and to search for the English comparative correlative construction \citep{Weissweiler2022}. The challenge here is that many previous methods for probing constructional knowledge \citep{Weissweiler2023a} are not applicable if we are looking for constructions that we do not yet know.

We take the mean embedding for each of the five hundred sentences in the dataset, averaged across the last two layers in the model. Because we are not concerned here with the contribution of specific layers, we use this averaged representation to capture the information available toward the final layer of the model. This use of pooled sentence embeddings is chosen to replicate the methods used in previous work \citep{Li2022b}.

For the sake of comparison, we include two embedding conditions: First, we use the raw embeddings, which of course capture both grammatical and non-grammatical information. These unaltered representations are called \textit{Direct Embeddings} in Tables \ref{tab:validation_embeddings} and \ref{tab:fp_embeddings}. Second, we create a grammar-focused embedding for each sentence that controls for lexical differences. This is done by also extracting the embedding for a shuffled version of the sentence, where word order is randomized. Given the sensitivity of English grammar to word order, this has the effect of removing some syntactic information, at least that which is not recoverable from lexical items \citep{Papadimitriou2022}. We then subtract this non-grammatical embedding from the original representation to create a representation which controls for lexical or topical information. These altered embeddings are called \textit{Grammar-Focused}. 

We then conduct the analysis across both sets of embeddings to ensure that any false positive constructions are not a confound of the lexical or topical attributes of the sentences. Finally, we follow this up with an introspective analysis of the results in order to search for additional possible confounds.

Our first step is to validate that these two sets of embeddings are able to correctly distinguish between the five true positive clause-level constructions in our data set. To do this, we train a logistic regression classifier with the goal of learning to distinguish between actual constructions. A high accuracy here would mean that these representations capture the grammatical generalizations that distinguish between these five constructions. These are true positives in the sense that linguists expect the grammatical representations of these constructions to be distinct. The results are shown by embedding type and construction type in Table \ref{tab:validation_embeddings}; these results are averaged across five-fold cross-validation. This level of accuracy validates that, if we were probing for actual true positive constructions, these methods would confirm the existence of those constructions for the model. Interestingly, the grammar-focused embeddings are worse at distinguishing constructions in all cases.

The next step is to develop a puppet hypothesis by trying to use these embeddings to find false positive constructions. The goal is find potential fake hypotheses that would also be confirmed by these same methods. For this we use k-means clustering to divide each set of sentences into two groups. This is a simple approach of creating a false distinction within each construction: according to the introspections of linguists, each category contains 100 examples of one and only one construction. We have divided these into two groups by clustering and then use a logistic regression classifier to test whether such a division would be confirmed as an actual constructional distinction. As before, a high accuracy means that the model confirms our analysis; the difference is that this is a decoy analysis. Note that we do not control for non-constructional factors like sentence length \citep{Weissweiler2023}, in part because we are searching for potential constructions rather than creating a test set for a hypothesized construction: we cannot manipulate these clusters. The introspection-based analysis at the end of this section, however, shows that there are no clear confounding factors in these two sets of sentences.\footnote{A quantitative analysis shows that these clusters could not be explained by factors like sentence length alone.}

\begin{table}
    \centering
    \begin{tabular}{|l|cc|}
        \hline
        ~ & \textit{Direct} & \textit{Grammar-} \\
        ~ & \textit{Embeddings} & \textit{Focused} \\
        \hline
         Intransitive & 0.99 & 0.99 \\
         Transitive (\textsc{np}) & 0.94 & 0.93 \\
         Transitive (\textsc{c}) & 0.96 & 0.99 \\
         Passive & NA & 0.99 \\
         Double Object & 0.97 & 0.97 \\
        \hline
    \end{tabular}
    \caption{Prediction accuracy by f-score for distinguishing between fake puppet constructions within each constructional categories. A high f-score means that the model hallucinates additional constructions that are not distinguished for humans. Reported numbers are the mean across 5-fold cross-validation.}
    \label{tab:fp_embeddings}
\end{table}

The results of this experiment are shown in Table \ref{tab:fp_embeddings}, across both actual constructions (rows) and the embedding conditions (columns). As before, \textit{Grammar-Focused} embeddings have been filtered to remove lexical or topical information, thus focusing more on structure. These results consistently show that the model makes distinctions between hallucinated constructions that do not exist for humans. In fact, the clarity of the hallucinated constructions (accuracy) is higher than of the true constructions. The only exception to this is the passive construction with direct embeddings; in this case, the clustering forms a single group and no classification probe is possible. In all other cases, this false hypothesis would have been confirmed.

It is possible, of course, that the model is more correct than human linguists in its analysis of constructions. Perhaps these new constructions discovered by the model are actually correct. Thus, we now ask: are these new constructional divisions linguistically motivated? To answer this question we undertake an introspection-based analysis of the clusters, starting with the Transitive (NP) sentences. Sentences from the model's first hallucinated construction are given in (16) and from the second hallucinated construction in (17).

~

\noindent(16a) You can change the color of a control. \\
\noindent(16b) In March 1613 he bought a gatehouse in the former Blackfriars priory. \\
\noindent(16c) His willful nature caused trouble throughout his life. \\

A comparison of these examples reveals that there is no grammatical distinction between these two sets of sentences. For instance, these sentences are paired by verb, with even the same sense of the same verb in the same clausal construction existing within both groups. And yet, if we had hypothesized that these sentences were instances of two separate constructions, the model would have confirmed our analysis with an f-score of 0.94.

~

\noindent(17a) You change the layout by moving the fields to predefined drop areas. \\
\noindent(17b) He bought a postcard of brilliant blue sea and dazzling white ruins. \\
\noindent(17c) They can cause property damage, create a mess, and produce unpleasant smells. \\

A further set of examples is given in (18) and (19), in this case representing two hallucinated constructions within the clausal argument transitive sentences. According to our introspection, these are all examples of a single construction. 

~

\noindent(18a) I realize that some were not signed by the artist.\\
\noindent(18b) We all know that John Kerry served in Vietnam. \\
\noindent(18c) I believe he must have waited among the gorse bushes through which the path winds. \\

As before, the examples of each hallucinated construction are aligned by verb, with (18a) comparable to (19a) and so on. And yet these do not form actual minimal pairs: each is still an instance of the same construction. These examples show that the groupings from the model, while robust, do not form a linguistically distinct set of utterances.

~

\noindent(19a) Nor did she realize that he wrote popular literature. \\
\noindent(19b) We all know that the market share of the railways has declined in recent years.\\
\noindent(19c) Once I returned to pick up my car, you can believe I spent quite a bit more time standing around waiting.\\

This section has conducted a probing experiment using contextual embeddings, first to distinguish between actual constructions and second to search for false positive constructions learned by the model. The high accuracy which validates the true positives is comparable to the high accuracy for the false positives. We then undertook a qualitative error analysis to determine if these new constructions had a legitimate but previously unknown linguistic basis. Our conclusion is that these do constitute false positive hallucinated constructions.\footnote{ Importantly, previous work which included additional NLI tasks along with the identification of constructions \citep{Weissweiler2023} would only have over-identified in the first task.} 

\section{Discussion and Conclusions}

The goal of this paper has been to investigate the possibility of false positive constructions in LLMs. If a question for computational syntax is whether these models are \textit{aware of} some syntactic structure, it is important to also search for constructions which the LLM is aware of incorrectly: hallucinated constructions that exist only for the model and not for humans.

This paper has shown that previous methods are inadequate for mapping the full syntactic knowledge of language models. Since we do not know how many such hallucinated constructions exist, there is a large piece missing in our understanding of how LLMs represent grammar. We can imagine two distinct scenarios: First, suppose that a human speaker of English knows 10k constructions and that an LLM knows 9k of those constructions. That would be a respectable a true positive rate of 90\%. But, second, suppose that the LLM knows 9k of the 10k actual constructions, but also an additional 20k hallucinated constructions. This would be a very different story, with a false positive rate exceeding the true positive rate. The problem is that previous work has been fundamentally unable to explore the possibility of false positive constructions.

From a linguistic perspective, this means that probing experiments should not yet be taken as evidence for a given linguistic analysis. Because even incorrect analyses can be confirmed in this way, we should not accept this form of evidence as support for linguistic theory itself. This means that Construction Grammar continues to struggle with falsifiability \citep{Cappelle2024}. In short, probing methods require minimal pairs which assume the existence of the construction to be tested. They cannot yet provide evidence for the existence of constructions themselves.

From a computational perspective, this means that we still do not know the full linguistic knowledge within LLMs because we have only looked for what we expected to find. What we do not know is the potentially vast store of incorrect constructional representations which have also been acquired by these models. Exploring the full range of such false positives remains a challenge for future work.

\textit{Minimal Pairs Cannot Be Formulated for Unknown False Positives}. Many of the core probing experiments in computational syntax are focused around minimal pairs which contrast specific phenomena: examples include active/passive alternations \citep{Leong2023}, dative/ditransitive alternations \citep{Hawkins2020}, and island effects \citep{Kobzeva2023}. A paradigm that relies on minimal pairs can be relatively confident in its true positives and false negatives. For instance, this kind of stimuli could be used to prove that a model does know island constraints or that a model does not know the restrained scope of the active/passive alternation. But, because hallucinated constructions are by definition unknown, it would never be possible to construct minimal pairs until they have been discovered. Thus, unless methods are developed to thoroughly search for syntactic hallucinations, we will never actually know the full range of syntactic knowledge of a model. 

\textit{Entrenchment and Exposure Are Specific to Individuals}. A further challenge is that, from a usage-based perspective, constructional representations are entrenched to various degrees. This means that a construction could be partially productive and it also means that the level of productivity could vary by individual \citep{Fonteyn2020, Dunn2021a} and by speech community \citep{hs11, d18b}. In short, usage-based theory makes claims about the grammars of specific groups who have had specific linguistic experiences. The challenge is that LLMs span speech communities and represent many different populations \citep{Dunn2024}. It is reasonable to say that speakers of American English have a given construction in a spoken register. But it is not reasonable to say that all speakers of English in all registers have that construction. Thus, another challenge is to determine what the benchmark population/register is for probing experiments. Does GPT-4 need to know all of the constructions of written American English? Of spoken Nigerian English? Of Indian English? Are these false positives actually entrenched constructions for other dialects? These are important questions to ask before we claim to understand the constructional knowledge which such models possess.

\textit{Can Computational Models Ever Tell Linguists Something They Did Not Know?} Previous work in constructional probing has focused always on confirming introspection-based analyses that linguists have already undertaken about phenomena that linguists already believed to be a part of the grammar. These methods as previously formulated can never discover new phenomena, and thus are unable to tell linguists something about grammar that they did not already expect to find. At the same time, as we have seen, these methods come with a confirmation bias which would incorrectly support hypotheses that we know have no basis.

One way forward is to develop methods for mapping the syntactic knowledge of LLMs which do not assume syntactic analyses from the start: what is the grammar that a model has learned, regardless of whether that grammar matches what linguists expect to find? As in this paper, such methods would do best to combine both explicit meta-linguistic knowledge with implicit behavioural knowledge. Further, since populations and individuals differ in their grammatical knowledge, it is important for such false positive probing experiments to also account for register and dialect.

\bibliographystyle{acl_natbib}
\bibliography{references}


\end{document}